\documentclass[11pt]{article}
\usepackage{acl2015}
\usepackage{times}
\usepackage{url}
\usepackage{latexsym}
\usepackage{enumitem}
\setlist{nosep} 
\usepackage{CJKutf8}
\usepackage{graphicx}
\usepackage{color}
\makeatletter
\newcommand{\@BIBLABEL}{\@emptybiblabel}
\newcommand{\@emptybiblabel}[1]{}
\makeatother
\usepackage[hidelinks]{hyperref}
\hypersetup{colorlinks=false}

\usepackage{amsmath}
\usepackage{caption}
\usepackage{subcaption}
\usepackage{float}

\newcommand{\parentinst}{\hat{c_i}}
\newcommand{\parentrole}{\hat{l_i}}
\newcommand{\amrprob}{P_{\mbox{\tiny AMR}}}
\newcommand{\roleprob}{P_{\mbox{\tiny Role}}}
\newcommand{\concept}[1]{\mbox{\texttt{#1}}}
\newcommand{\rolelabel}[1]{\mbox{\textbf{#1}}}

\title{Using Syntax-Based Machine Translation to Parse English into Abstract Meaning Representation}

\author{Michael Pust, Ulf Hermjakob, Kevin Knight, Daniel Marcu, Jonathan May \\
  Information Sciences Institute \\
  Computer Science Department\\
  University of Southern California \\
  {\tt \{pust, ulf, knight, marcu, jonmay\}@isi.edu} \\}

\date{}

\begin{document}

\maketitle

\begin{abstract}
We present a parser for Abstract Meaning Representation (AMR).  We treat English-to-AMR conversion within the framework of string-to-tree, syntax-based machine translation (SBMT).  To make this work, we transform the AMR structure into a form suitable for the mechanics of SBMT and useful for modeling. We introduce an AMR-specific language model and add data and features drawn from semantic resources. Our resulting AMR parser improves upon state-of-the-art results by 7 Smatch points.
\end{abstract}

\section{Introduction}

\begin{figure}
\begin{subfigure}[b]{.5\textwidth}
\centering
\includegraphics[width=2in]{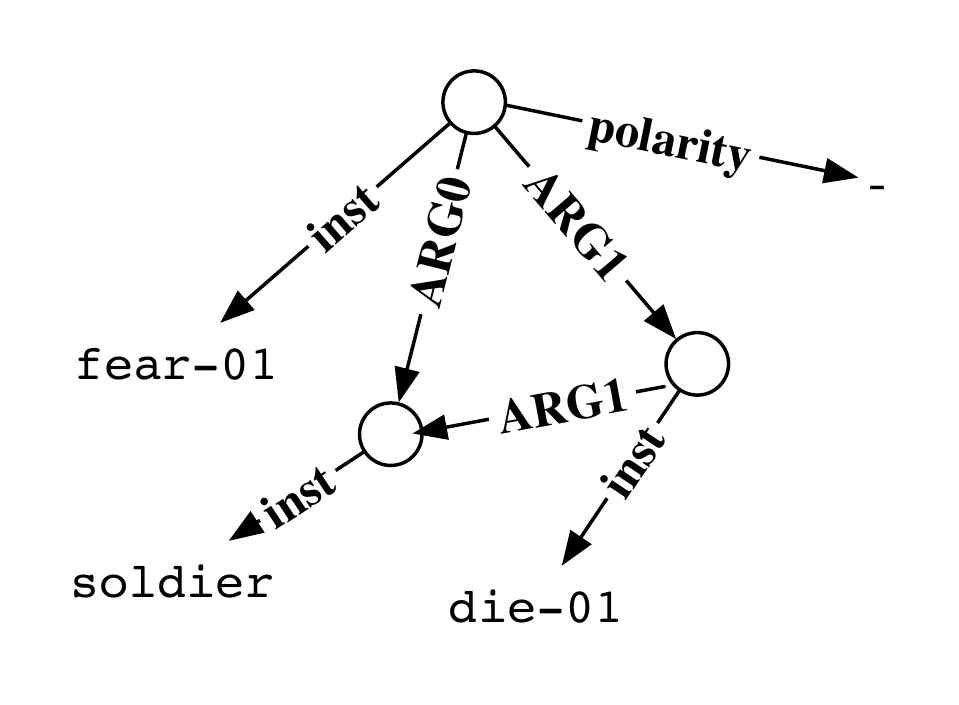}
\end{subfigure}

\begin{subfigure}[b]{.5\textwidth}
\centering
\begin{tabular}{c}
The soldier was not afraid of dying. \\
The soldier was not afraid to die. \\
The soldier did not fear death. \\
\end{tabular}
\end{subfigure}
\caption{An Abstract Meaning Representation (AMR) with several English renderings.}
\label{amr}
\end{figure}

Abstract Meaning Representation (AMR) is a compact, readable, whole-sentence semantic annotation~\cite{banarescu-EtAl:2013:LAW7-ID}.  It includes entity identification and typing, PropBank semantic roles \cite{Kingsbury02fromtreebank}, individual entities playing multiple roles, as well as treatments of modality, negation, etc.  AMR abstracts in numerous ways, e.g., by assigning the same conceptual structure to {\em fear} (v), {\em fear} (n), and {\em afraid} (adj).  Figure~\ref{amr} gives an example of an AMR with several English renderings.

In this paper, we automatically learn how to parse English into AMR, by exploiting a publicly available corpus of more than 10,000 English/AMR pairs.\footnote{LDC Catalog number 2014T12}  

The AMR parsing problem bears a strong formal resemblance to syntax-based machine translation (SBMT) of the string-to-tree variety, as shown in Figure~\ref{similarities}.  Because of this, it is appealing to consider whether we can apply the substantial body of techniques already invented for SBMT.\footnote{See e.g. the related work section of \newcite{huck-hoang-koehn:2014:W14-33}.}


\begin{figure}[h]
\begin{center}
\small
\begin{tabular}{|l|l|l|} \hline
& SBMT & AMR parsing \\ \hline
Source & flat string & flat string \\ \hline
Target & nested structure & nested structure \\ \hline
\end{tabular}
\end{center}
\caption{Similarities between AMR parsing and syntax-based machine translation (SBMT).}
\label{similarities}
\end{figure}

However, there are also some major differences, as shown in Figure~\ref{differences}.  Therefore, applying SBMT to AMR parsing requires novel representations and techniques, which we develop in this paper.  Some of our key ideas include:

\begin{figure}[t]
\begin{center}
\small
\begin{tabular}{|p{.6in}|p{.825in}|p{.825in}|} \hline
& SBMT & AMR parsing \\ \hline
Target & tree & graph \\ \hline
Nodes & labeled & unlabeled \\ \hline
Edges & unlabeled & labeled \\ \hline
Alignments & words to leaves & words to leaves \\ 
           &                 & + words to edges \\ \hline
Children & ordered & unordered \\ \hline
Accuracy Metric & {\sc Bleu} \cite{Papineni:2002:BMA:1073083.1073135} & Smatch \cite{cai-knight:2013:Short}   \\ 
\hline

\end{tabular}
\end{center}
\caption{Differences between AMR parsing and SBMT.}
\label{differences}
\end{figure}

\begin{enumerate}
\item Introducing an AMR-equivalent representation that is suitable for string-to-tree SBMT rule extraction and decoding.
\item Proposing a target-side reordering technique that takes unique advantage of the fact that child nodes in AMR are unordered.
\item Introducing an AMR-specific language model.
\item Developing tuning methods that maximize Smatch \cite{cai-knight:2013:Short}.  
\item Integrating 
several semantic knowledge sources into the AMR parsing task.
\end{enumerate}

\noindent
These key ideas lead to state-of-the-art AMR parsing results.  We next describe our baseline SBMT system, and then we adapt it to AMR parsing.

\section{Syntax-Based Machine Translation}

Our baseline SBMT system proceeds as follows.  Given a corpus of (source string, target tree, source-target word alignment) sentence translation training tuples and a corpus of (source, target, score) sentence translation tuning tuples:

\begin{enumerate}
\item \textbf{Rule extraction:} A grammar of string-to-tree rules is induced from training tuples using the GHKM algorithm \cite{galley-EtAl:2004:HLTNAACL,galley-EtAl:2006:COLACL}.
\item \textbf{Local feature calculation:} Statistical and indicator features, as described by \newcite{chiang-knight-wang:2009:NAACLHLT09}, are calculated over the rule grammar.
\item \textbf{Language model calculation:} A Kneser-Ney-interpolated 5-gram language model \cite{Chen:1996:ESS:981863.981904} is learned from the yield of the target training trees.
\item \textbf{Decoding:} A beamed bottom-up chart decoder
calculates the optimal derivations given a source string and feature parameter set.
\item \textbf{Tuning:} Feature parameters are optimized using the MIRA learning approach \cite{chiang-knight-wang:2009:NAACLHLT09} to maximize the objective, typically \textsc{Bleu} \cite{Papineni:2002:BMA:1073083.1073135},  associated with a tuning corpus.
\end{enumerate}

We  initially use this system with no modifications and pretend that English--AMR is a language pair indistinct from any other.

\begin{figure*}[th]

\begin{subfigure}[b]{0.3\textwidth}
\includegraphics[width=2in]{{soldierorig}.pdf}
\caption{The original AMR}
\label{fig:orig}
\end{subfigure}
\hfill
\begin{subfigure}[b]{0.3\textwidth}
\includegraphics[width=2in]{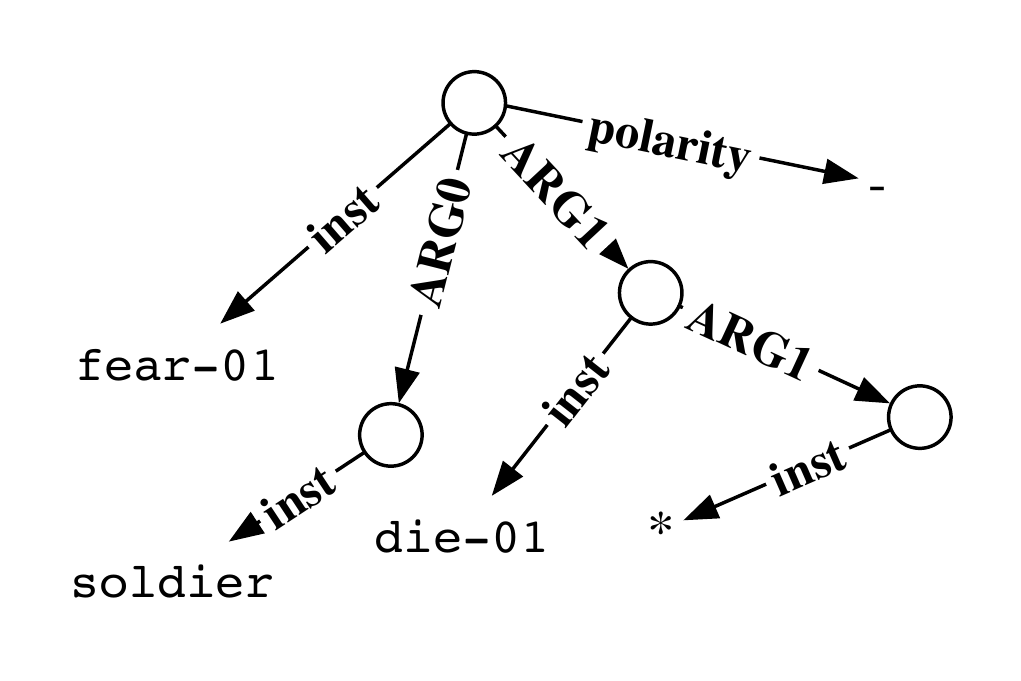}
\caption{Disconnecting multiple parents of (a)}
\label{fig:disconnect}
\end{subfigure}
\hfill
\begin{subfigure}[b]{0.3\textwidth}
\includegraphics[width=2in]{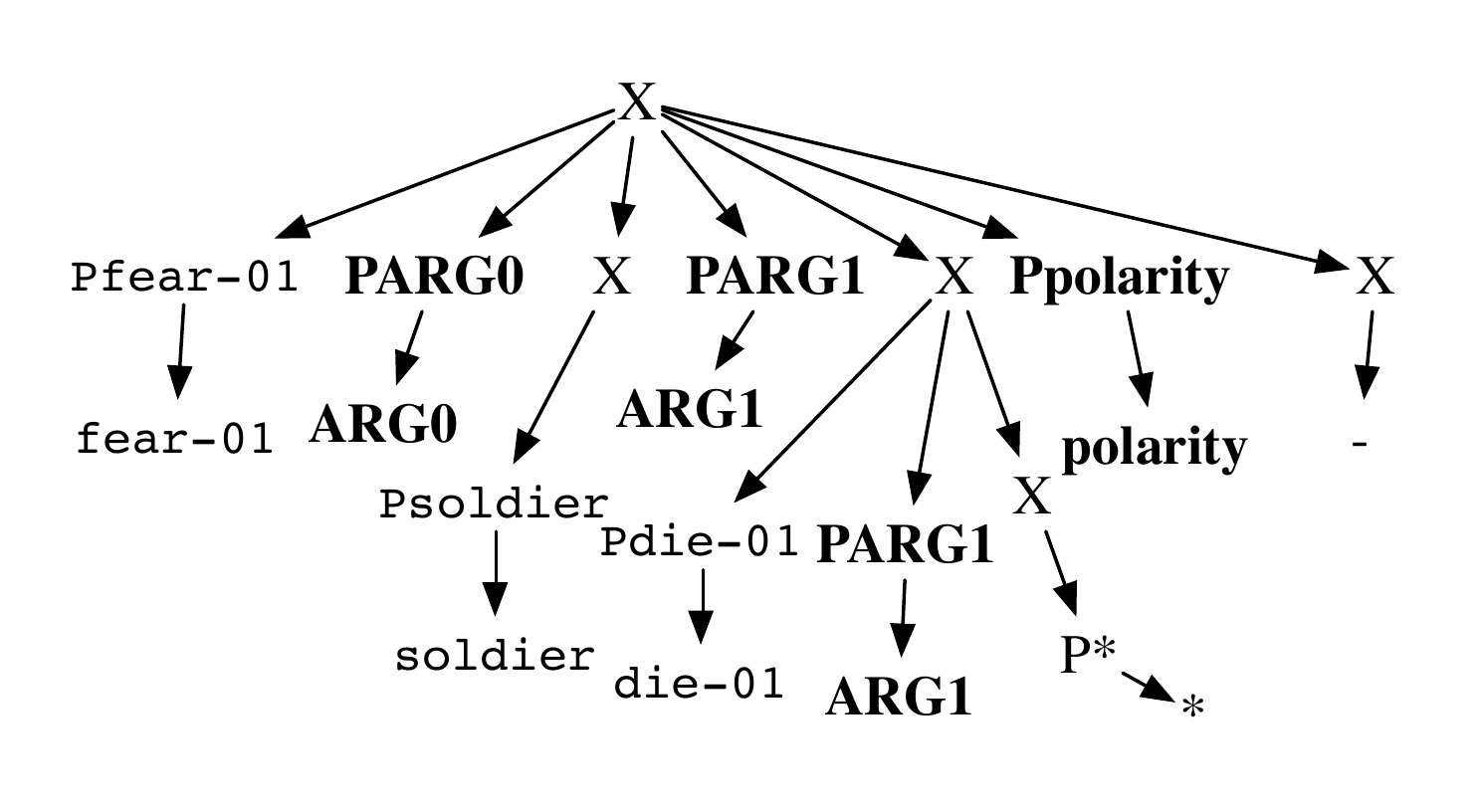}
\caption{Edge labels of (b) pushed to leaves, preterminals added}
\label{fig:flattree}
\end{subfigure}

\begin{subfigure}[b]{0.3\textwidth}
\includegraphics[width=2in]{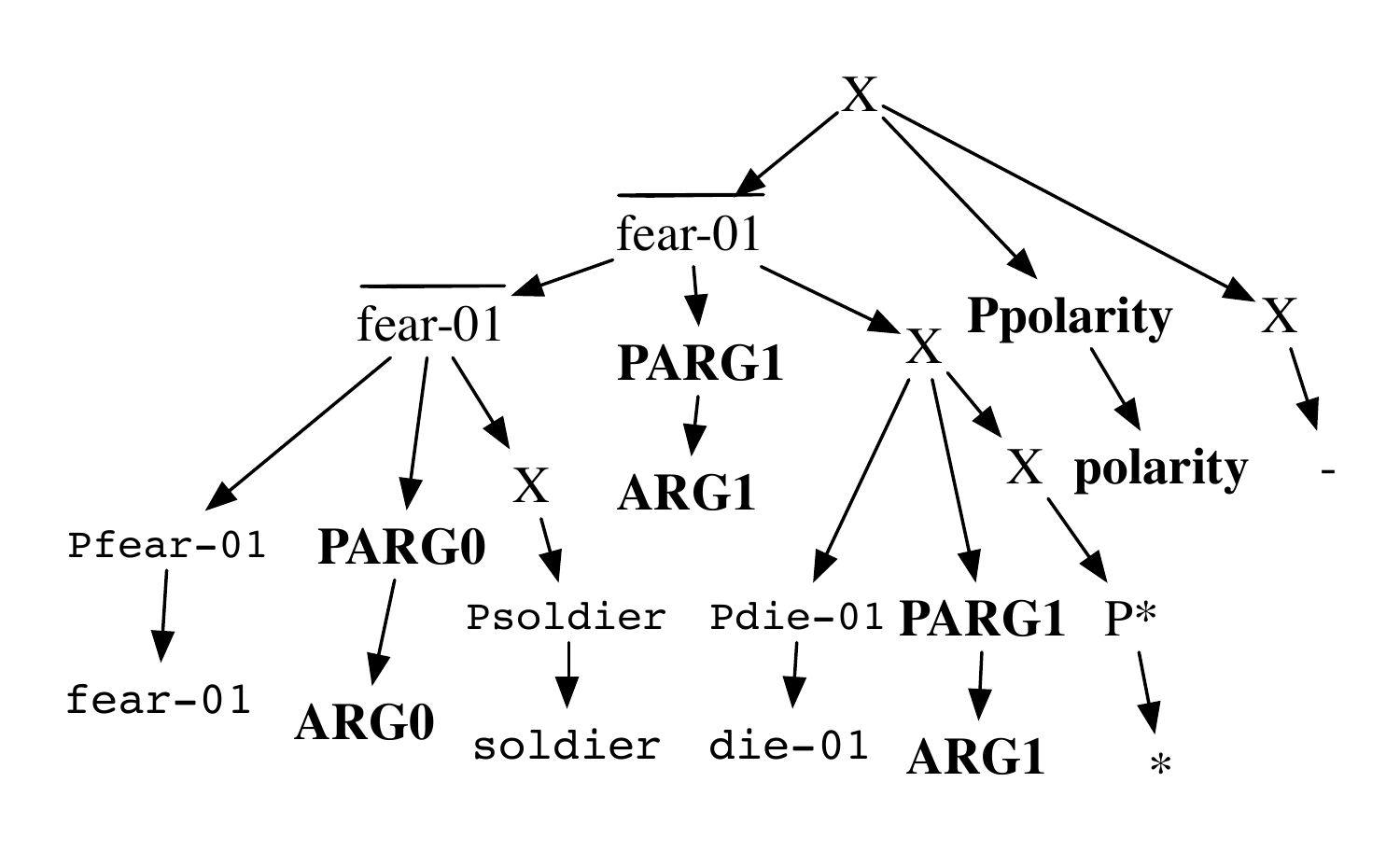}
\caption{Restructuring (c) with concept labels as intermediates}
\label{fig:concrestr}
\end{subfigure}
\hfill
\begin{subfigure}[b]{0.3\textwidth}
\includegraphics[width=2in]{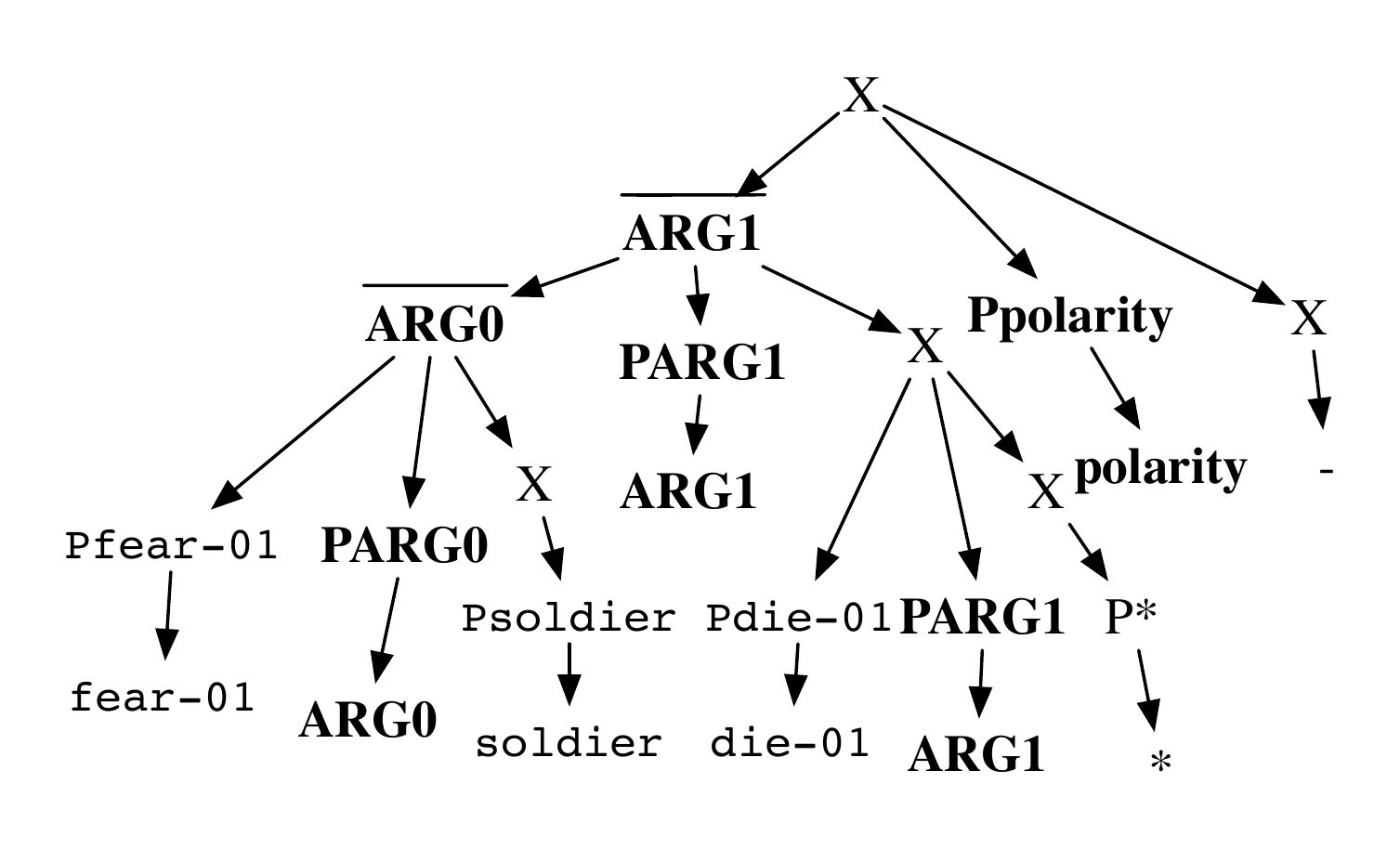}
\caption{Restructuring (c) with role labels as intermediates}
\label{fig:rolerestr}
\end{subfigure}
\hfill
\begin{subfigure}[b]{0.3\textwidth}
\includegraphics[width=2in]{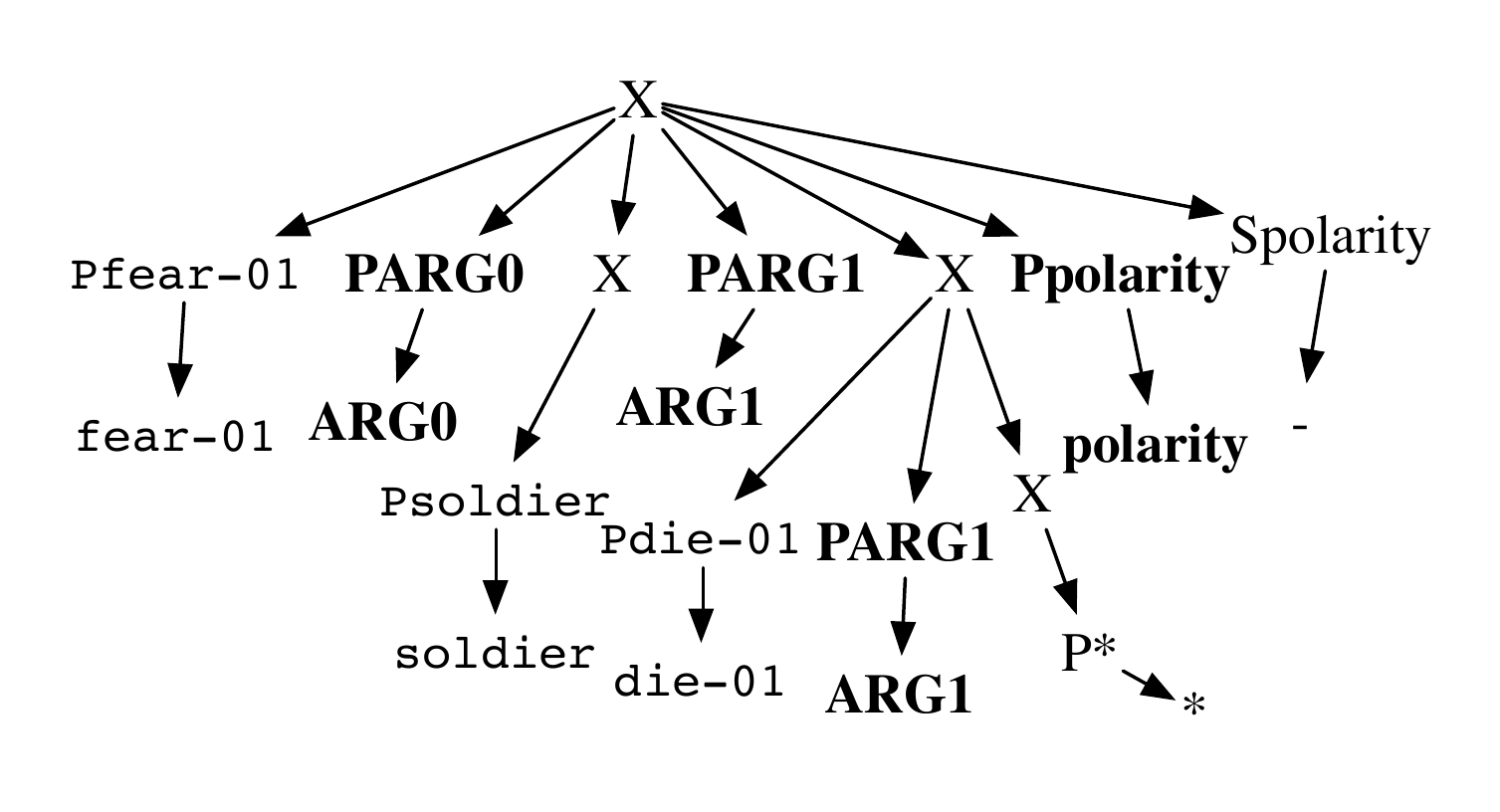}
\caption{String preterminal relabeling of (c)}
\label{fig:nterm}
\end{subfigure}

\begin{subfigure}[b]{0.3\textwidth}
\includegraphics[width=2in]{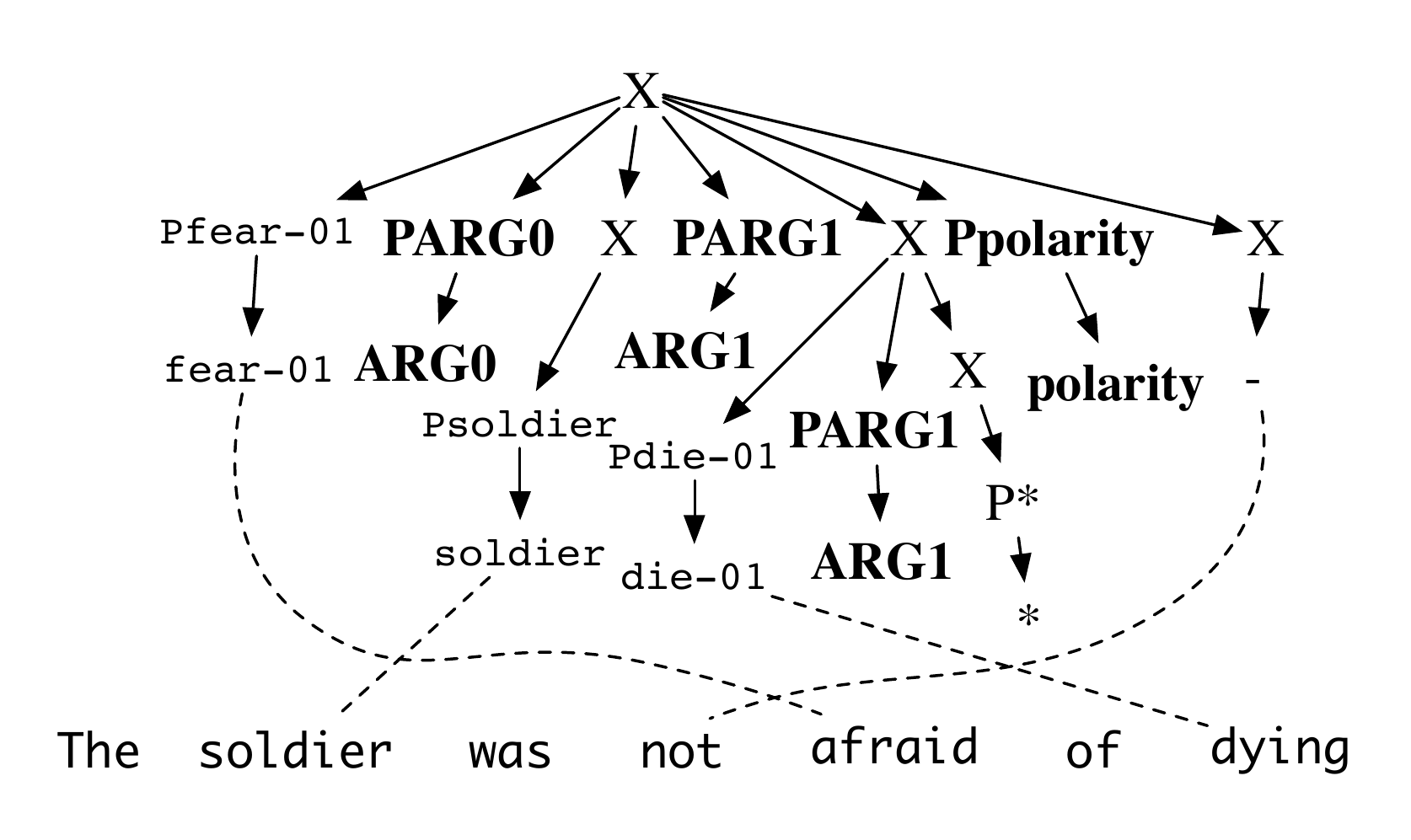}
\caption{Original alignment of English to (c)}
\label{fig:origalign}
\end{subfigure}
\hfill
\begin{subfigure}[b]{0.3\textwidth}
\includegraphics[width=2in]{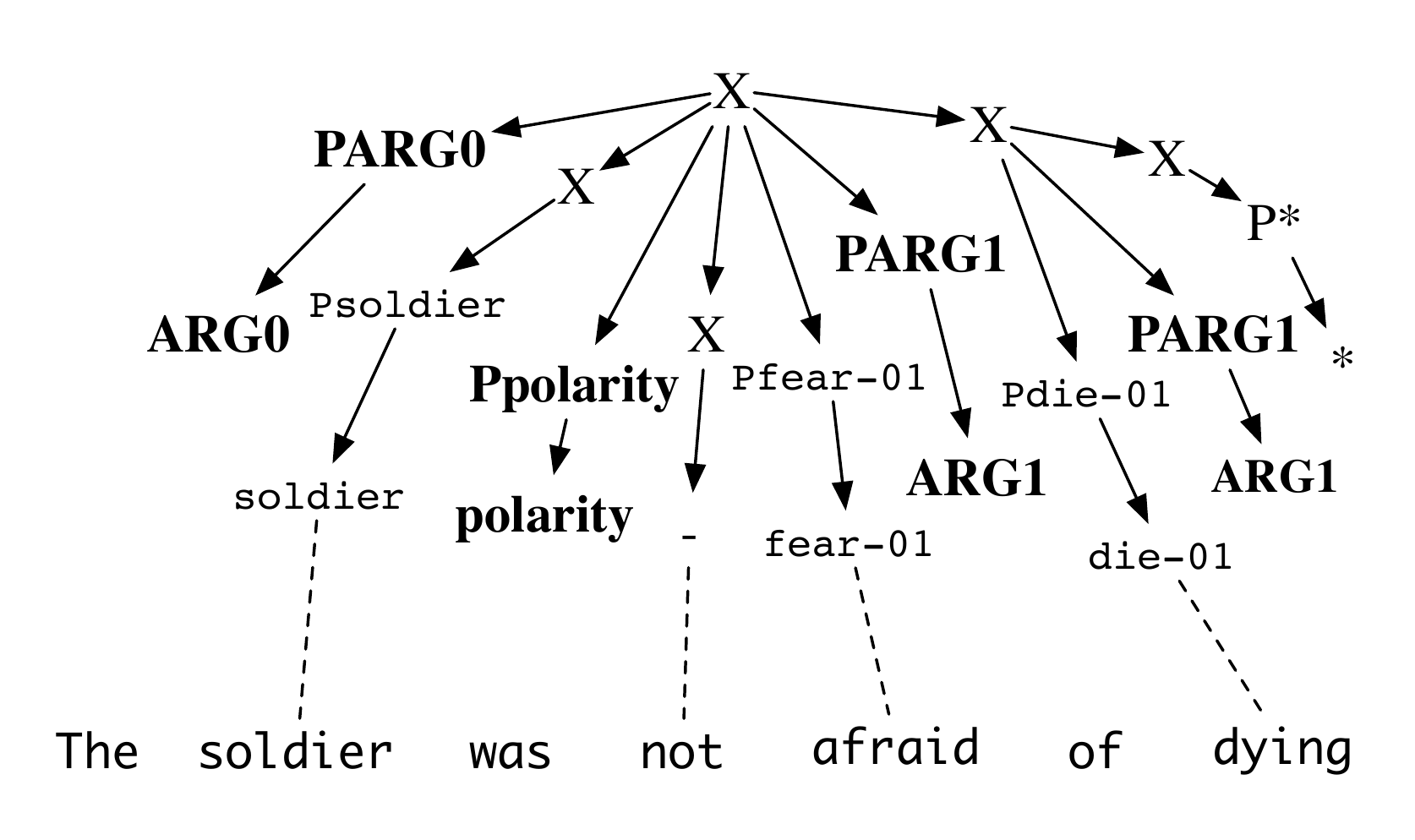}
\caption{After reordering of (g)}
\label{fig:reorder}
\end{subfigure}
\hfill
\begin{subfigure}[b]{0.3\textwidth}
\includegraphics[width=2in]{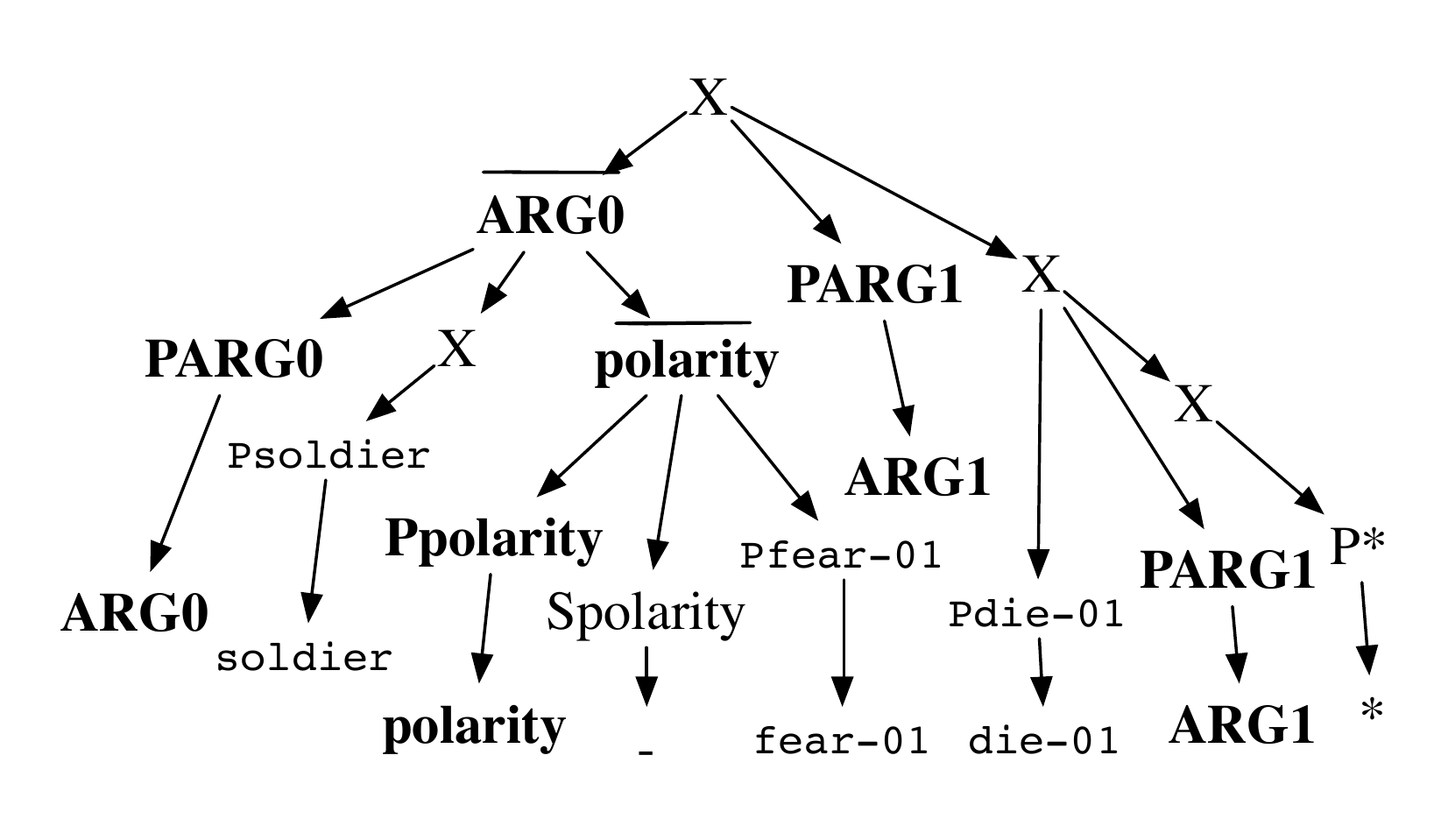}
\caption{Restructured, relabeled, and reordered tree: (e), (f), and (h)}
\label{fig:finished}
\end{subfigure}

\caption{Transformation of AMR into tree structure that is acceptable to GHKM \protect\cite{galley-EtAl:2004:HLTNAACL,galley-EtAl:2006:COLACL} rule extraction and yields good performance.}
\label{fig:maketree}
\end{figure*}

\section{Data and Comparisons}
\label{sec:data}

We use English--AMR data from the AMR 1.0 corpus, LDC Catalog number 2014T12. In contrast to narrow-domain data sources that are often used in work related to semantic parsing \cite{Price:1990:ESL:116580.116612,Zelle:1996:UIL:239549,AAAI04ws-pillar}, the AMR corpus covers a broad range of news and web forum data. We use the training, development, and test splits specified in the AMR corpus (Table~\ref{tab:data}). The training set is used for rule extraction, language modeling, and statistical rule feature calculation. The development set is used both for parameter optimization and qualitatively for hill-climbing. The test set is held out blind for evaluation. We preprocess the English with a simple rule-based tokenizer and, except where noted, lowercase all data. We obtain English--AMR alignments by using the unsupervised alignment approach of \newcite{pourdamghani-EtAl:2014:EMNLP2014}. All parsing results reported in this work are obtained with the Smatch 1.0 software \cite{cai-knight:2013:Short}. We compare our results to those of \newcite{flanigan-etal:ACL2014} on the AMR 1.0 data splits; we run that work's JAMR software according to the provided instructions.\footnote{\url{https://github.com/jflanigan/jamr}}

\begin{table}
\centering
\begin{tabular}{r|ll}
Corpus & Lines & Tokens \\
\hline
Training & 10,313 & 218,021 \\
Development & 1,368 & 29,484 \\
Test & 1,371 & 30,263 \\
\hline
\end{tabular}
\caption{Data splits of AMR 1.0, used in this work. Tokens are English, after tokenization.}
\label{tab:data}
\end{table}
\section{AMR Transformations}
\label{sec:rrr}

In this section we discuss various transformations to our AMR data. Initially, we concern ourselves with converting AMR into a form that is amenable to GHKM rule extraction and string to tree decoding. We then turn to structural transformations designed to improve system performance. Figure~\ref{fig:maketree} progressively shows all the transformations described in this section; the example we follow is shown in its original form in Figure~\ref{fig:orig}.

\subsection{Massaging AMRs into Syntax-Style Trees}
\label{sec:flat}

The relationships in AMR form a Directed Acyclic Graph (DAG), but GHKM requires a tree, so we must begin our transformations by discarding some information. We arbitrarily disconnect all but a single parent from each node (see Figure~\ref{fig:disconnect}). This is the only lossy modification we make to our AMR data. As multi-parent relationships occur approximately once per training sentence, this is indeed a regrettable loss. We nevertheless make this modification, since it allows us to use the rest of our string-to-tree tools.

AMR also contains labeled edges, unlike the constituent parse trees we are used to working with in SBMT. These labeled edges have informative content and we would like to use the alignment procedure of 
\newcite{pourdamghani-EtAl:2014:EMNLP2014}, which aligns words to edges as well as to terminal nodes. So that our AMR trees are compatible with both our desired alignment approach and our desired rule extraction approach, we propagate edge labels to terminals via the following procedure:

\begin{enumerate}
\item For each node $n$ in the AMR tree we create a corresponding node $m$ with the all-purpose symbol `X' in the SBMT-like tree. Outgoing edges from $n$ come in two flavors: \textit{concept edges}, labeled `inst', which connect $n$ to a terminal \textit{concept} such as \texttt{fear-01}, and \textit{role edges}, which have a variety of labels such as \textbf{ARG0} and \textbf{name}, and connect $n$ to another instance or to a \textit{string}.\footnote{In Figure~\ref{fig:maketree} the negative polarity marker `-' is a string. Disconnected referents labeled `*' are treated as AMR instances with no roles.} A node has one instance edge and zero or more role edges. We consider each type of edge separately.
\item For each outgoing role edge we insert two unlabeled edges into the corresponding transformation; the first is an edge from $m$ to a terminal bearing the original edge's role label (a so-called \textit{role label edge}), and the second (a \textit{role filler edge}) connects $m$ to the transformation of the original edge's target node, which we process recursively. String targets of a role receive an `X' preterminal to be consistent with the form of role filler edges.
\item For the outgoing concept edge we insert an unlabeled edge connecting $m$ and the concept. It is unambiguous to determine which of $m$'s edges is the concept edge and which edges constitute role label edges and their corresponding role filler edges, as long as paired label and filler edges are adjacent.
\item Since SBMT expects trees with preterminals, we simply replicate the label identities of concepts and role labels, adding a marker (`P' in Figure~\ref{fig:maketree}) to distinguish preterminals.
\end{enumerate}

The complete transformation can be seen in Figure~\ref{fig:flattree}. Apart from multiple parent ancestry, the original AMR can be reconstructed deterministically from this SBMT-compliant rewrite.

\begin{figure*}[tH]

\begin{subfigure}[b]{0.35\textwidth}
\centering
\includegraphics[width=2.5in]{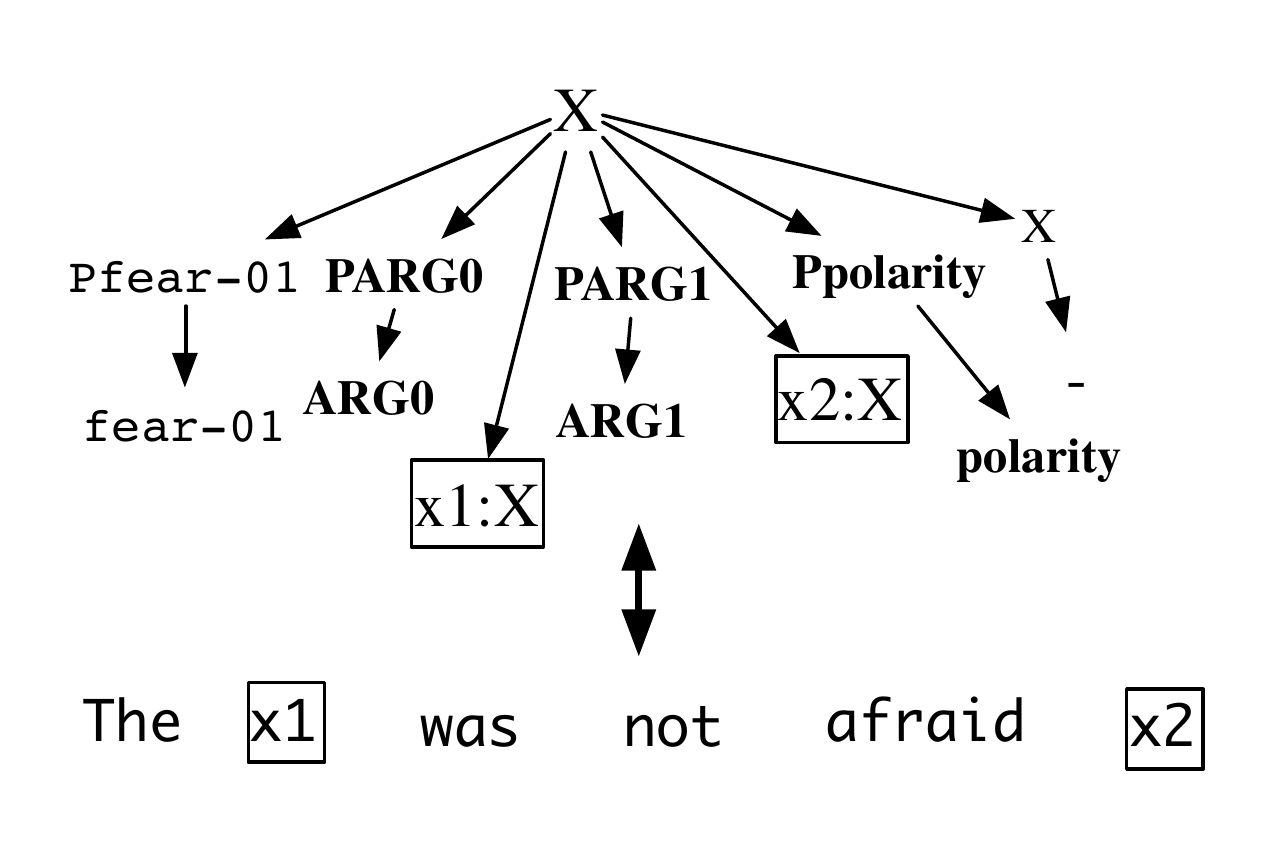}
\caption{A rule extracted from the AMR tree of Figure \ref{fig:flattree}. All roles seen in training must be used.}
\label{fig:flatrule}
\end{subfigure}
\hfill
\begin{subfigure}[b]{0.25\textwidth}
\centering
\includegraphics[width=1.75in]{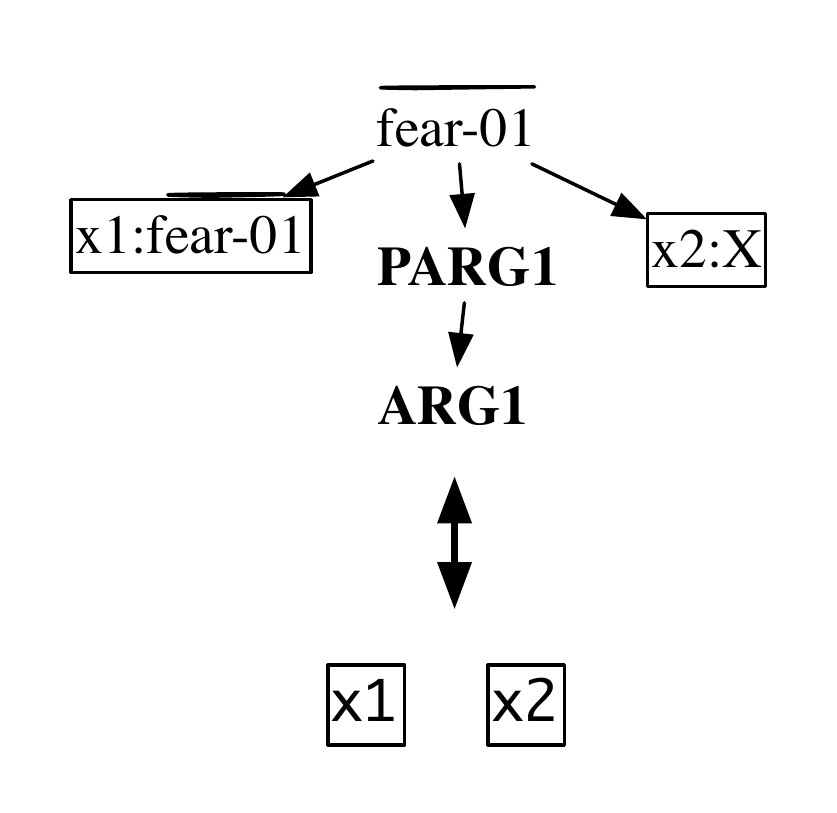}
\caption{A rule from the AMR tree of Figure \ref{fig:concrestr}.
Many nearly identical rules of this type are extracted, and this rule can be used multiple times in a single derivation.
}
\label{fig:conceptrule}
\end{subfigure}
\hfill
\begin{subfigure}[b]{0.3\textwidth}
\centering
\includegraphics[width=1.75in]{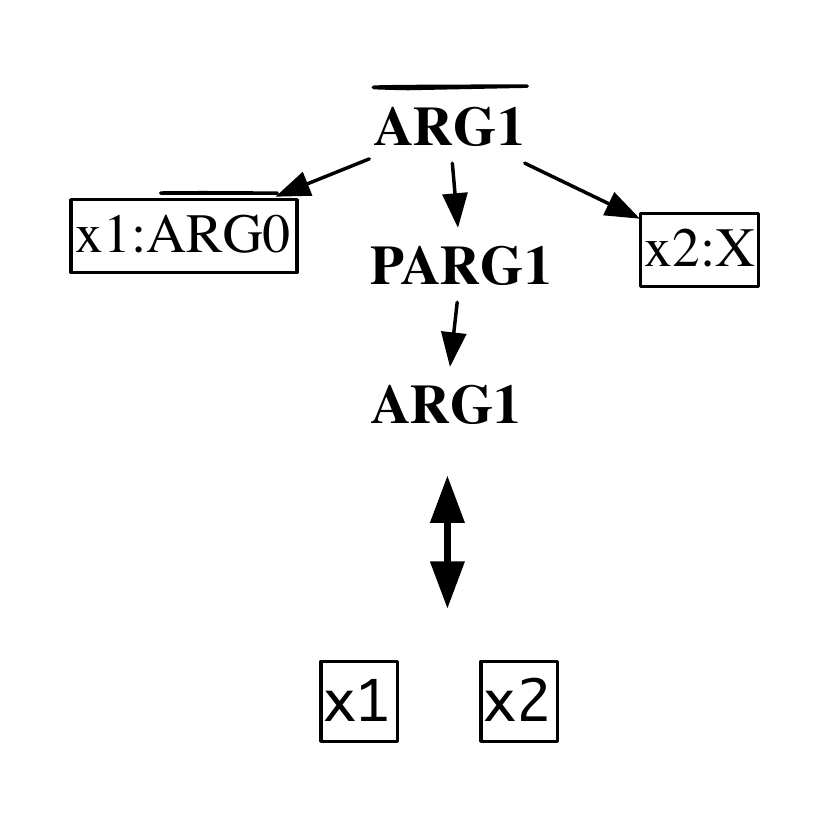}
\caption{A rule from the AMR tree of Figure \ref{fig:rolerestr}. 
This rule can be used independent of the concept context it was extracted from and multiple reuse is discouraged.
}
\label{fig:rolerule}
\end{subfigure}
\caption{Impact of restructuring on rule extraction}
\label{fig:rules}
\end{figure*}

\subsection{Tree Restructuring}
\label{sec:rr}

While the transformation in Figure~\ref{fig:flattree} is acceptable to GHKM, and hence an entire end-to-end AMR parser may now be built with SBMT tools, we do not believe the resulting parser will exhibit very good performance. The trees we are learning on are exceedingly flat, and thus yield rules that do not generalize sufficiently. Rules produced from the top of the tree in Figure~\ref{fig:flattree}, such as that in Figure~\ref{fig:flatrule}, are only appropriate for cases where \texttt{fear-01} has exactly three roles: \textbf{ARG0} (agent), \textbf{ARG1} (patient), and \textbf{polarity}.

We follow the lead of \newcite{rexing10}, who in turn were influenced by similar approaches in monolingual parsing \cite{collins:1997:ACL,Charniak:2000:MP:974305.974323}, and re-structure trees at nodes with more than three children (i.e. instances with more than one role), to allow generalization of flat structures.

However, our trees are unlike syntactic constituent trees in that they do not have labeled nonterminal nodes, so we have no natural choice of an intermediate (``bar'') label. We must choose a meaningful label to characterize an instance and its roles. We initially choose the concept label, resulting in trees like that in Figure~\ref{fig:concrestr}. However, this attempt at re-structuring yields rules like that in Figure~\ref{fig:conceptrule}, which are general in form but are tied to the concept context in which they were extracted. This leads to many redundant rules and blows up the nonterminal vocabulary size to approximately 8,000, the size of the concept vocabulary. Furthermore, the rules elicited by this procedure encourage undesirable behavior such as the immediate juxtaposition of two rules generating \textbf{ARG1}. We next consider restructuring with the immediately dominant role labels, resulting in trees like that in Figure~\ref{fig:rolerestr} and rules like that in Figure~\ref{fig:rolerule}. This approach leads to more useful rules with fewer undesirable properties. 

\subsection{Tree Relabeling}
\label{sec:rl} 
 
AMR strings have an effective preterminal label of `X,' which allows them to compete with full AMR instances at decode time. However, whether or not a role is filled by a string or an instance is highly dependent on the kind of role being filled. The \textbf{polarity} and \textbf{mode} roles, for instance, are nearly always filled by strings, but \textbf{ARG0} and \textbf{ARG1} are always filled by instances. The \textbf{quant} role, which is used for representation of numerical quantities, can be filled by an instance (e.g. for approximate quantities such as `about 3') or a string. To capture this behavior we relabel string preterminals of the tree with labels indicating role identity and string subsumption. This relabeling, replaces, for example, one `X' preterminal in Figure~\ref{fig:flattree} with ``Spolarity,''  as shown in Figure~\ref{fig:nterm}.

\subsection{Tree Reordering}
\label{sec:ro}

Finally, let us consider the alignments between English AMR. As is known in SBMT, non-monotone alignments can lead to large, unwieldy rules and in general make decoding more difficult. While this is often an unavoidable fact of life when trying to translate between two languages with different syntactic behavior, it is an entirely artificial phenomenon in English--AMR. AMR is an \textit{unordered} representation, yet in order to use an SBMT infrastructure we must declare an order of the AMR tree. This means we are free to choose whatever order is most convenient to us, as long as we keep role label edges immediately adjacent to their corresponding role filler edges to preserve conversion back to the edge-labeled AMR form. We thus choose the order that is as close as possible to English yet still preserves these constraints. We use a simple greedy bottom-up approach that permutes the children of each internal node of the unrestructured tree so as to minimize crossings. This leads to a 79\% overall reduction in crossings and is exemplified in Figure~\ref{fig:origalign} (before) and Figure~\ref{fig:reorder} (after). We may then restructure our trees, as described above, in an instance-outward manner. The final restructured, relabeled, and reordered tree is shown in Figure~\ref{fig:finished}. 

\section{AMR Language Models}
\label{sec:amrlm}

We now turn to language models of AMRs, which help us prefer reasonable target structures over unreasonable ones.

Our first language model is unintuitively simple---we pretend there is a language called {\em AMRese} that consists of yields of our restructured AMRs.  An example AMRese string from Figure~\ref{fig:finished} is `\textbf{ARG0} \texttt{soldier} \texttt{polarity} - \texttt{fear-01} \textbf{ARG1} \texttt{die-01} \textbf{ARG1} *.'  We then build a standard n-gram model for AMRese.

It also seems sensible to judge the correctness of an AMR by calculating the empirical probability of the concepts and their relations to each other. This is the motivation behind the following model of an AMR:\footnote{This model is only defined over AMRs that can be represented as trees, and not over all AMRs. Since tree AMRs are a prerequisite of our system we did not yet investigate whether this model could be sufficiently generalized.}

We define an AMR instance $i = (c, R)$, where $c$ is a \textit{concept} and $R$ is a set of \textit{roles}. 
We define an AMR role $r = (l, i)$, where $l$ is a role label, and $i$ is an AMR instance labeled $l$. 
For an AMR instance $i$ let $\parentinst$ be the concept of $i$'s parent instance, and $\parentrole$ be the label of the role that $i$ fills with respect to its parent. 
We also define the special instance and role labels ROOT and STOP. Then, we define $\amrprob(i|\parentrole, \parentinst)$, the conditional probability of AMR instance $i$ given its ancestry as:

\vspace{-.2in}
\begin{align*}
&\amrprob(i=(c, R)| \parentrole, \parentinst)  =  P(c|\parentrole, \parentinst) \times \\
&                                \prod\limits_{r \in R} \roleprob(r|c)  \times P(\mbox{STOP}|c) \\
\mbox{where} &\\
&\roleprob(r = (l, i)| c)  =  P(l|c) \times  \amrprob(i|l, c) \\
\end{align*}
\vspace{-.2in}

We define $P(c|\parentrole, \parentinst)$, $P(l|c)$, and $P(\mbox{STOP}|c)$ as empirical conditional probabilities, Witten-Bell interpolated \cite{WittenBell91} to lower-order models by progressively discarding context from the right.\footnote{That is, $P(c|\parentrole, \parentinst)$ is interpolated with $P(c|\parentrole)$ and then $P(c)$.} 
We model exactly one STOP event per instance.
We define the probability of a full-sentence AMR $i$ as $\amrprob(i | \mbox{ROOT})$ where ROOT in this case serves as both parent concept and role label.

As an example, the instance associated with concept \texttt{die-01} in Figure~\ref{fig:disconnect} has $\parentrole =$ \textbf{ARG1} and $\parentinst = $ \texttt{fear-01}, so we may score it as:

\vspace{-.2in}
\begin{align*} 
& P(\concept{die-01}|\rolelabel{ARG1}, \concept{fear-01}) \times \\ 
& P(\rolelabel{ARG1}|\concept{die-01}) \times \\
& P(\mbox{STOP}|\concept{die-01}) \times \\
& P(\mbox{*}|\rolelabel{ARG1}, \concept{die-01})
\end{align*}
\vspace{-.2in}

We add the AMR LM to our system as a feature alongside the AMRese n-gram LM.

\begin{table*}[t]

\centering
\begin{tabular}{|r|l|ll|}
\hline
System & Section & Tune & Test \\
\hline
flat trees & \ref{sec:flat} & 51.6 & 49.9 \\
concept restructuring & \ref{sec:rr} & 57.2 & 55.3 \\
role restructuring (rr) & \ref{sec:rr} & \textbf{60.8} & \textbf{58.6} \\
rr + string preterminal relabeling (rl) & \ref{sec:rl} &  \textbf{61.3} & \textbf{59.7} \\
rr + rl + reordering (ro) & \ref{sec:ro} & \textbf{61.7} & \textbf{59.7} \\
rr + rl + ro + AMR LM & \ref{sec:amrlm} & \textbf{62.3} & \textbf{60.6} \\
rr + rl + ro + AMR LM + date/number/name rules (dn) & \ref{sec:dn} & \textbf{63.3} & \textbf{61.3} \\
rr + rl + ro + AMR LM + dn + semantic categories (sc) & \ref{sec:sc} & \textbf{66.2} & \textbf{64.3} \\
rr + rl + ro + AMR LM + dn + sc, rule-based alignments & \ref{sec:ua} & \textbf{67.1} & \textbf{65.3} \\ 
rr + rl + ro + AMR LM + dn + sc, rule-based + unsupervised alignments & \ref{sec:ua} & \textbf{67.7} & \textbf{65.8} \\ 

\hline
JAMR \cite{flanigan-etal:ACL2014} & \ref{sec:data} & 58.8 & 58.2 \\
\hline
\end{tabular}
\caption{AMR parsing Smatch scores for the experiments in this work. We provide a cross-reference to the section of this paper that  describes each of the evaluated systems. Entries in \textbf{bold} are improvements over the previous state of the art. Human inter-annotator Smatch performance is in the 79-83 range \protect\cite{cai-knight:2013:Short}. }
\label{tab:mainresults}
\end{table*}

\section{Adding External Semantic Resources}

While we are engaged in the task of semantic parsing, we have not yet discussed the use of any semantic resources. In this section we rectify that omission.

\subsection{Rules from Numerical Quantities and Named Entities}
\label{sec:dn}

While the majority of string-to-tree rules in SBMT systems are extracted from annotated data, it is common practice to dynamically generate rules to handle the translation of dates and numerical quantities, as these follow common patterns and are easily detected at decode-time. We follow this practice here, and additionally detect person names at decode-time  using the Stanford Named Entity Recognizer \cite{finkel-grenager-manning:2005:ACL}. We use cased, tokenized source data to build the decode-time rules. We add indicator features to these rules so that our tuning methods can decide how favorable the resources are. We leave as future work the incorporation of named-entity rules for other classes, since most available named-entity recognition beyond person names is at a granularity level that is incompatible with AMR (e.g. we can recognize `Location' but not distinguish between `City' and `Country').

\subsection{Hierarchical Semantic Categories}
\label{sec:sc}

\begin{figure}
\centering
\includegraphics[width=3in]{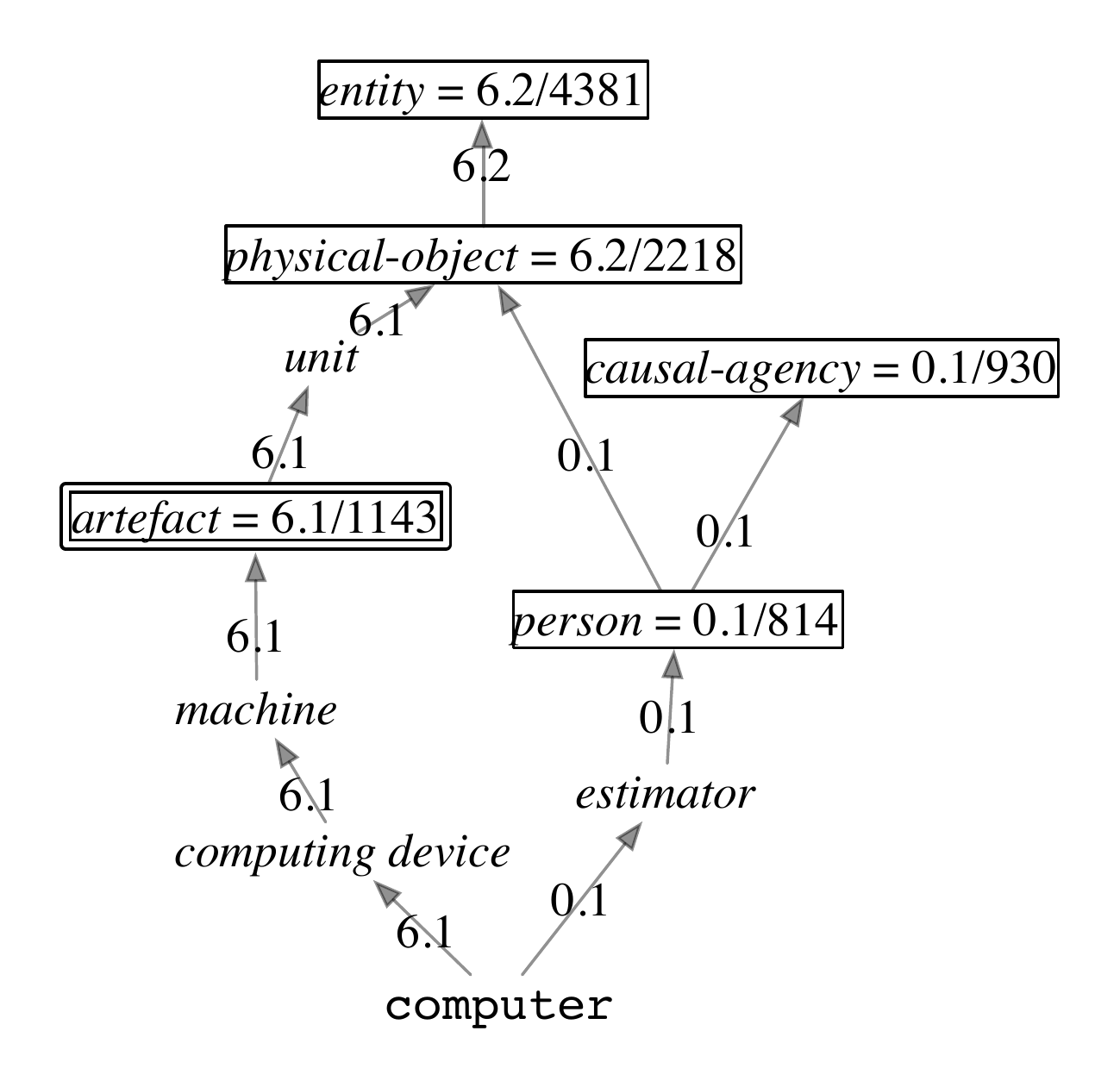}
\caption{WordNet hierarchy for \texttt{computer}. Pre-selected salient WordNet categories are boxed. Smoothed sense counts are propagated up the hierarchy and re-combined at join points. Scores are calculated by dividing propagated sense count by count of the category's prevalence over the set of AMR concepts. The double box indicates the selection of \textit{artefact} as the category label for \texttt{computer}.}
\label{fig:mapping}
\end{figure}

In order to further generalize our rules, we modify our training data AMRs once more, this time replacing the identity preterminals over concepts with preterminals designed to enhance the applicability of our rules in semantically similar contexts. For each concept $c$ expressed in AMR, we consult WordNet \cite{wordnet} and a curated set of gazetteers and vocabulary lists to identify a hierarchy of increasingly general semantic categories that describe the concept. So as not to be overwhelmed by the many fine-grained distinctions present in WordNet, we pre-select around 100 salient semantic categories from the WordNet ontology. When traversing the WordNet hierarchy, we propagate a smoothed count\footnote{We use very simple smoothing, and add 0.1 to the provided example counts.} of the number of examples seen per concept sense\footnote{Since WordNet senses do not correspond directly to PropBank or AMR senses, we simply use a lexical match and must consider all observed senses for that match.}, combining counts when paths meet. For each selected semantic category $s$ encountered in the traversal, we calculate a weight by dividing the propagated example count for $c$ at $s$ by the frequency $s$ was proposed over all AMR concepts. We then assign $c$ to the highest scoring semantic category $s$. An example calculation for the concept \texttt{computer} is shown in Figure~\ref{fig:mapping}.

\begin{figure}
\centering
\includegraphics[width=3in]{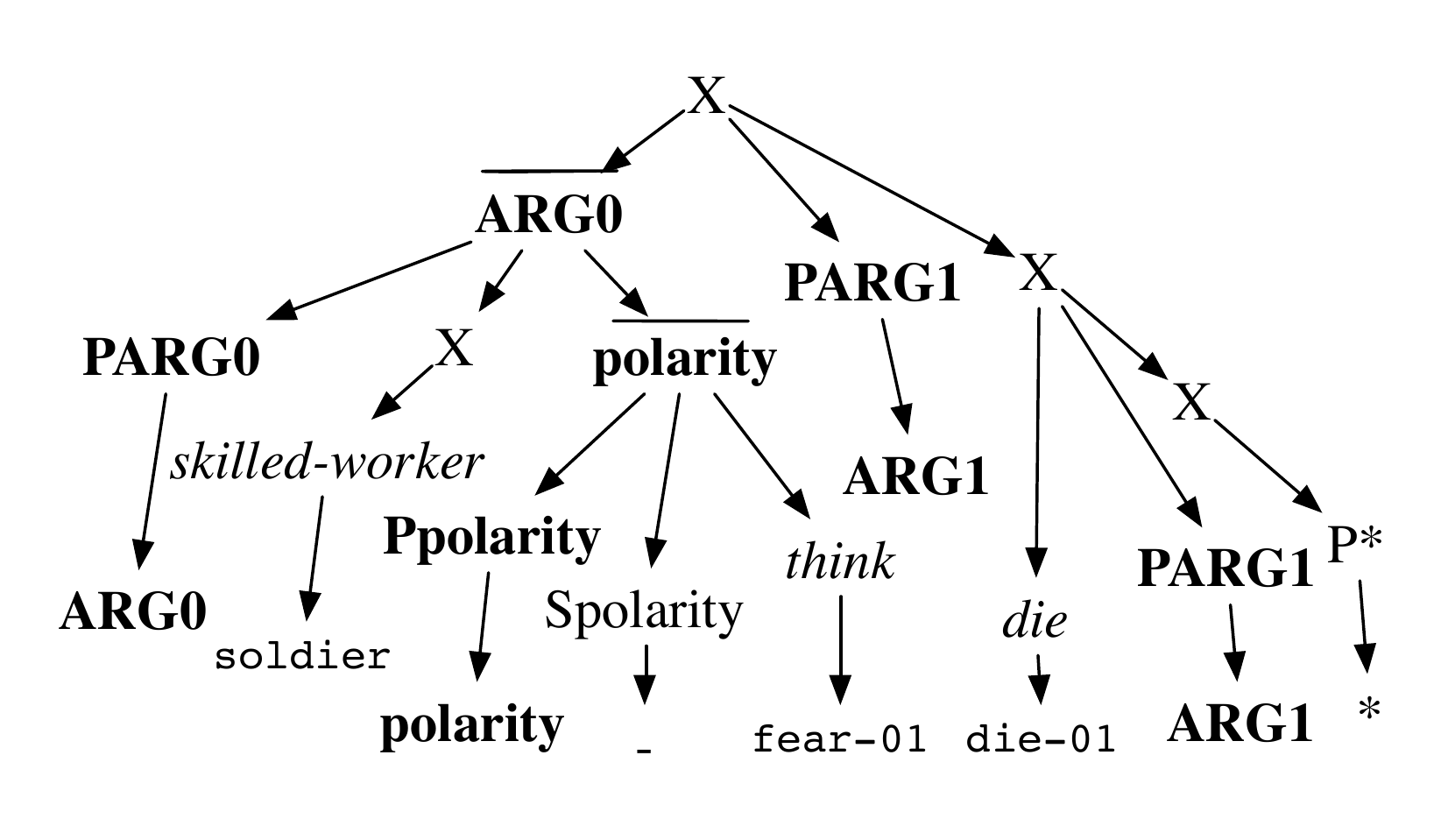}
\caption{Final modification of the AMR data; semantically clustered preterminal labels are added to concepts.}
\label{fig:semlabels}
\end{figure}

We apply semantic categories to our data as replacements for identity preterminals of concepts. This leads to more general, more widely-applicable rules. For example,  with this transformation, we can parse correctly not only contexts in which ``soldiers die'', but also contexts in which other kinds of ``skilled workers die''. Figure \ref{fig:semlabels} shows the addition of semantic preterminals to the tree from Figure \ref{fig:finished}. We also incorporate semantic categories into the AMR LM. For concept $c$, let $s_c$ be the semantic category of $c$. Then we reformulate $\amrprob(i|\parentrole, \parentinst)$ as:

\vspace{-.2in}
\begin{align*}
&\amrprob(i=(c, R)| \parentrole, \parentinst)  = \\
& P(s_c | \parentrole, s_{\parentinst}, \parentinst) \times P(c|s_c, \parentrole, s_{\parentinst}, \parentinst) \times \\
& \prod\limits_{r \in R} \roleprob(r|c) \times P(\mbox{STOP}|s_c, c) \\
 \mbox{where}& \\
& \roleprob(r = (l, i)|c) = P(l|s_c, c) \times \amrprob(i|l, c) \\
\end{align*}
\vspace{-.4in}

\subsection{Semantically informed Rule-based Alignments}
\label{sec:ua}

For our final incorporation of semantic resources we revisit the English-to-AMR alignments used to extract rules. As an alternative to the unsupervised approach of \newcite{pourdamghani-EtAl:2014:EMNLP2014}, we build alignments by taking a linguistically-aware, supervised heuristic approach  to alignment:

First, we generate a large number of potential links between English and AMR. We attempt to link English and AMR tokens after conversion through resources such as a morphological analyzer, a list of 3,235 pertainym pairs (e.g. adj-`gubernatorial' $\to$ noun-`governor'), a list of 2,444 adverb/adjective pairs (e.g. `humbly' $\to$ `humble'), a list of 2,076 negative polarity pairs (e.g. `illegal' $\to$ `legal'), and a list of 2,794 known English-AMR transformational relationships (e.g. `asleep' $\to$  \texttt{sleep-01}, `advertiser' $\to$ \texttt{person :ARG0-of advertise-01}, `Greenwich Mean Time' $\to$ \texttt{GMT}). These links are then culled based on context and AMR structure. For example, in the sentence ``The big fish ate the little fish,'' initially both English `fish' are aligned to both AMR `fish.' However, based on the context of `big' and `little' the spurious links are removed.

In our experiments we explore both replacing the unsupervised alignments of \newcite{pourdamghani-EtAl:2014:EMNLP2014} with these alignments and concatenating the two alignment sets together, essentially doubling the size of the training corpus. Because the different alignments yield different target-side tree reorderings, it is necessary to build separate 5-gram AMRese language models.\footnote{The AMR LM is insensitive to reordering so we do not need to vary it when varying alignments.} When using both alignment sets together, we also use both AMRese language models simultaneously.

\begin{figure}

\includegraphics[width=3in]{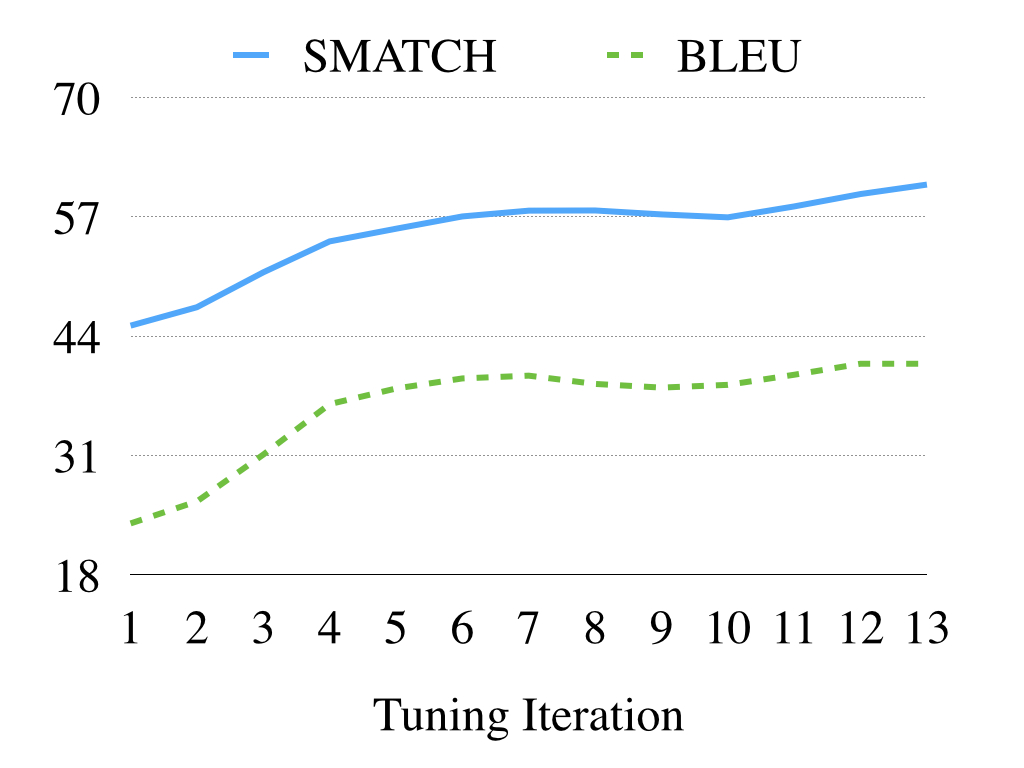}

\caption{{\sc Bleu} of AMRese and Smatch correlate closely when tuning.}
\label{fig:correlation}
\end{figure}

\section{Tuning}

We would like to tune our feature weights to maximize Smatch directly.  However, a very convenient alternative is to compare the AMRese yields of candidate AMR parses to those of reference AMRese strings, using a \textsc{Bleu} objective and forest-based MIRA \cite{chiang-knight-wang:2009:NAACLHLT09}. Figure~\ref{fig:correlation} shows that MIRA tuning with \textsc{Bleu} over AMRese tracks closely with Smatch. Note that, for experiments using reordered AMR trees, this requires obtaining similarly permuted reference tuning AMRese and hence requires alignments on the development corpus. When using unsupervised alignments we may simply run inference on the trained alignment model to obtain development alignments.\footnote{In practice, since the corpus is not too large and the released implementation by \newcite{pourdamghani-EtAl:2014:EMNLP2014} is based on the multi-threaded mGIZA++ \cite{Gao08parallelimplementations}, we align twice, once with just the training data, and once with training and development data. We use the training-only run for rule extraction and only the development part of the training+dev for tuning.} The rule-based aligner runs one sentence at a time and can be employed on the development corpus. When using both sets of alignments, each approach's AMRese is used as a development reference (i.e. each development sentence has two possible reference translations).

\section{Results}

Our AMR parser's performance is shown in Table~\ref{tab:mainresults}. We progressively show the incremental improvements and compare to the current state of the art system of \newcite{flanigan-etal:ACL2014}. Purely transforming AMR data into a form that is compatible with the SBMT pipeline yields suboptimal results, but by adding role-based restructuring, relabeling, and reordering, as described in Section \ref{sec:rrr} we are able to surpass \newcite{flanigan-etal:ACL2014}. Adding an AMR LM and semantic resources increases scores further. Rule-based alignments are an improvement upon unsupervised alignments, but concatenating the two alignments yields the best results.

\section{Related Work}

The only prior AMR parsing work we are aware of is that of \newcite{flanigan-etal:ACL2014}. In that work, multiple discriminatively trained models are used to identify individual concept instances and then a minimum spanning tree algorithm connects the concepts.

Several other recent works have used a machine translation approach to semantic parsing, but all have been applied to domain data that is much narrower and an order of magnitude smaller than that of AMR, primarily the Geoquery corpus \cite{Zelle:1996:LPD:1864519.1864543}. The WASP system of \newcite{wong-mooney:2006:HLT-NAACL06-Main} uses hierarchical SMT techniques and does not apply semantic-specific improvements. \newcite{andreas-vlachos-clark:2013:Short} use phrase-based and hierarchical SMT techniques on Geoquery.  Like this work, they perform a transformation of the input semantic representation so that it is amenable to use in an existing machine translation system. However, they are unable to reach the state of the art in performance. \newcite{conf/aaai/LiLS13} directly address GHKM's word-to-terminal alignment requirement by extending that algorithm to handle word-to-node alignment.

Earlier work on using machine translation techniques for semantic parsing includes that of  \newcite{DBLP:conf/interspeech/PapineniRW97}. That work applies the IBM machine translation models \cite{Brown93themathematics} to English paired with a non-structural formal language of air travel queries. In a similar vein, \newcite{Macherey01naturallanguage} use IBM models and their Alignment Template approach to analyze a relatively large corpus of German train scheduling inquiries. The formal language they generate is not structural and has a vocabulary of less than 30 types; it may thus be seen as an instance of semantic role labeling rather than semantic representation parsing.

Our SBMT system is grounded in the theory of tree transducers, which have also been applied to the task of semantic parsing by Jones et al. \shortcite{BevanJonesMarkJohnsonSharonGoldwater:2011:ALTA2011,jones-johnson-goldwater:2012:ACL2012}.

Semantic parsing in general and AMR parsing specifically can be considered a subsumption of multiple semantic resolution sub-tasks, such as named entity recognition \cite{ner-sekine2007}, semantic role labeling \cite{Gildea:2002:ALS:643092.643093}, word sense disambiguation \cite{Navigli:2009:WSD:1459352.1459355} and relation finding \cite{bach2007review}.

\section{Conclusion}

By restructuring our AMRs we are able to convert a sophisticated SBMT engine into a baseline semantic parser with little additional effort. By further restructuring our data to appropriately model the behavior we want to capture we are able to rapidly achieve state-of-the-art results. Finally, by incorporating novel language models and external semantic resources, we are able to increase quality even more. This is  not the last word on AMR parsing, as fortunately, machine translation technology provides more low-hanging fruit to pursue. 

\section*{Acknowledgments}
This work was supported by DARPA contracts FA8750-13-2-0045, HR0011-12-C-0014, and W911NF-14-1-0364. 

\bibliographystyle{acl}
\bibliography{semparse}

\end{document}